\documentclass[pdflatex,sn-mathphys-num,iicol]{sn-jnl}
\usepackage[utf8]{inputenc}

\usepackage{graphicx}%
\usepackage{multirow}%
\usepackage{amsmath,amssymb,amsfonts}%
\usepackage{amsthm}%
\usepackage{mathrsfs}%
\usepackage[title]{appendix}%
\usepackage{xcolor}%
\usepackage{textcomp}%
\usepackage{manyfoot}%
\usepackage{booktabs}%
\usepackage{algorithm}%
\usepackage{algorithmicx}%
\usepackage{algpseudocode}%
\usepackage{listings}%

\usepackage{array}  
\usepackage{siunitx}  
\usepackage[normalem]{ulem} 
\usepackage{enumitem}

\usepackage{soul} 

\newcommand{\reduce}[1]{}

\newcommand{\highlight}[1]{#1}

\begin{document}

\title[Online Learning for Vibration Suppression in Physical Robot Interaction using Power Tools]{Online Learning for Vibration Suppression in Physical 

Robot Interaction using Power Tools}

\author*[1]{\fnm{Gokhan} \sur{Solak}}\email{gokhan.solak@iit.it}

\author[1]{\fnm{Arash} \sur{Ajoudani}}\email{arash.ajoudani@iit.it}

\affil*[1]{\orgdiv{HRI$^2$ Lab}, \orgname{Italian Institute of Technology (IIT)}, \orgaddress{\street{Via S. Quirico, 19D}, \city{Genoa}, \postcode{16163}, \country{Italy}}}



\abstract{

Vibration suppression is an important capability for collaborative robots deployed in challenging environments such as construction sites. 
We study the active suppression of vibration caused by external sources such as power tools. 
We adopt the band-limited multiple Fourier linear combiner (BMFLC) algorithm to learn the vibration online and counter it by feedforward force control.
We propose the \textit{damped BMFLC} method, extending BMFLC with a novel adaptive step-size approach that improves the convergence time and noise resistance. 
Our logistic function-based damping mechanism reduces the effect of noise and enables larger learning rates.
We evaluate our method on extensive simulation experiments with realistic time-varying multi-frequency vibration and real-world physical interaction experiments. 
The simulation experiments show that our method improves the suppression rate in comparison to the original BMFLC and its recursive least squares and Kalman filter-based extensions. Furthermore, our method is far more efficient than the latter two. 
\highlight{We further validate the effectiveness of our method in real-world polishing experiments.} 
A supplementary video is available at \href{https://youtu.be/ms6m-6JyVAI}{https://youtu.be/ms6m-6JyVAI}.
}


\maketitle

\section{Introduction}

Strong and persistent vibration is harmful for both human and machine health. 
In humans, long exposure to vibrating power tools may induce health problems, such as the hand-arm vibration syndrome \cite{heaver2011hand}. 
Instead in machines, vibration undermines the precision in control applications and may lead to mechanical wear \cite{collette2011review,yan2021bio}. 
For these reasons, vibration suppression is an important capability for employing collaborative robots in new environments such as construction sites \cite{brosque2020human} where the vibration is a common phenomenon. 

\begin{figure}[t]
\centering
  \includegraphics[width=0.97\columnwidth]{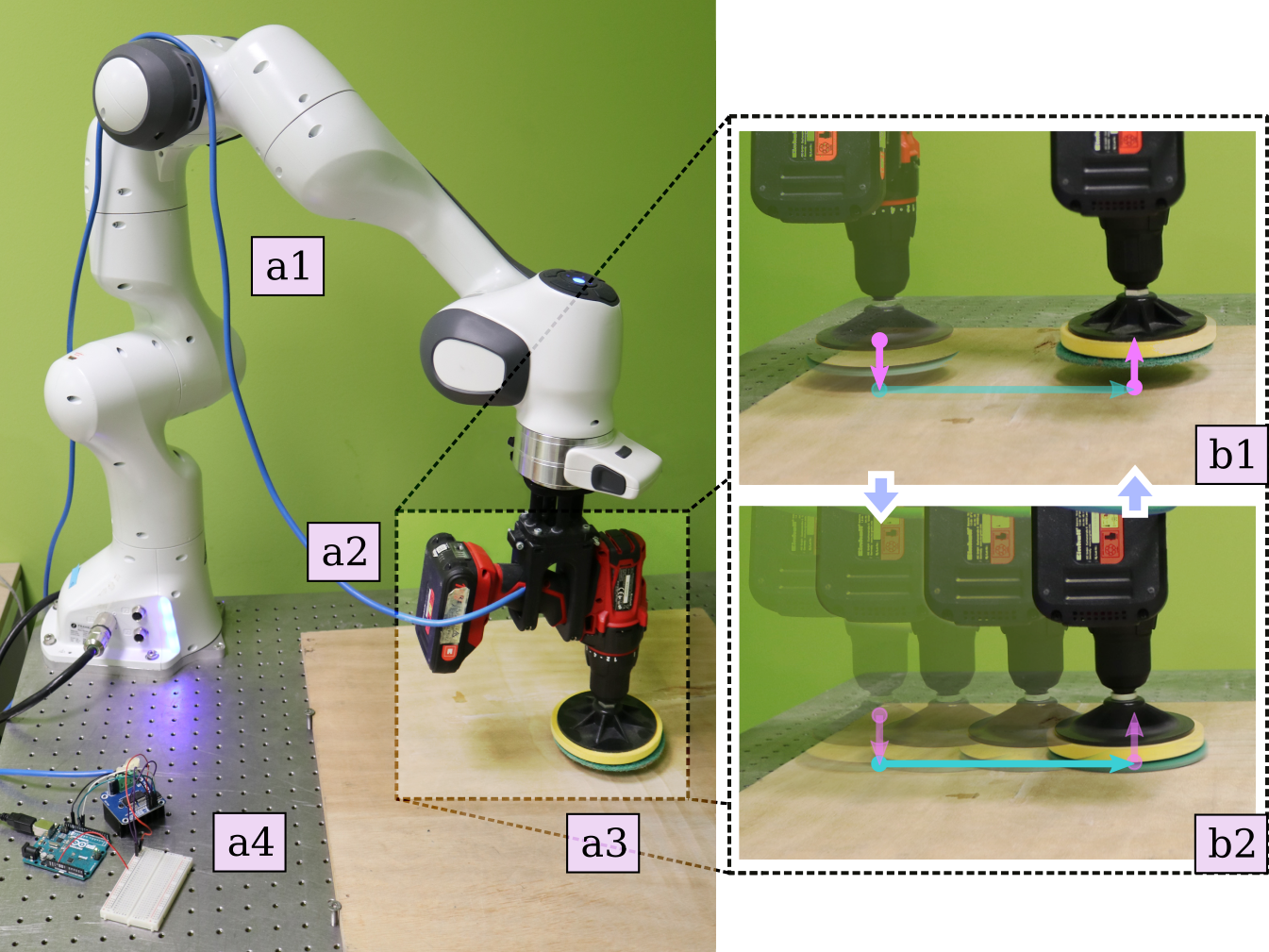} 
  \caption{We test our vibration suppression method on a 7-DoF Panda robot arm (a1) carrying a modified hand-held drill (a2) with an abrasive head to polish a wood block (a3). The rotation speed of the drill is controlled via an Arduino-USB interface (a4). The polishing motion is shown on the right. The experiment starts and ends without contact (b1). The polisher makes physical contact and moves \textit{20 cm} without breaking contact (b2), then lifted again. } 
  \label{fig:experiment-setup}%
\end{figure}

This work builds on our preliminary results \cite{solak2023online}, in which we studied the feedforward vibration suppression in human-robot collaboration (HRC). 
The main outcome of our study was that feedforward force control increases the vibration suppression performance while maintaining a compliant impedance profile, in comparison to the variable impedance control (VIC) approach which was previously used in HRC literature for dealing with vibrations \cite{okunev2012human,campeau2016time}. 
We successfully applied the BMFLC algorithm \cite{veluvolu2007bandlimited} in our high-dof robotic arm for learning and suppressing the vibration online. In this work, we extend both our theoretical approach and experiments. 

Firstly, we propose the \textit{damped BMFLC}, a new adaptive step-size method to improve the BMFLC learning rate and robustness to noise. The original Fourier linear combiner (FLC) and BMFLC algorithms use the least mean squares (LMS) method for optimising the model weights~\cite{vaz1989adaptive,veluvolu2007bandlimited}. 
LMS is a widely used, versatile optimisation approach, however, it converges at a slow rate with a fixed step-size. 
Our preliminary experiments showed that increasing the step-size above a certain level causes the algorithm to learn the noise, adding controller-induced vibrations.  
We modify the update function with a damping mechanism that penalises noise, which is indicated by smaller weights of the model. 
Our method allows both a fast changing and stable relationship between the model weights and the learning rate.

Secondly, we demonstrate the generalisation capacity of the proposed method in dealing with time-varying multi-frequency vibrations. 
In real-world applications, the vibration frequency may change due to the unstructured and varying interactions with the environment and it can consist of multiple frequency components. 
Accordingly, we synthesise complex vibrating motions and initially test our method in extensive simulation experiments.
We compare our method to the original BMFLC algorithm and its prevalent extensions which replace LMS with the Recursive least squares (RLS) and Kalman filter methods. 

Our simulation experiments show that our method repeatedly achieves the top performance over vibrations of varying characteristics. Moreover, it is much more computationally efficient than the RLS and Kalman filter alternatives.
Then, we run extended simulation experiments to explore the limits of our method. This gives valuable insight on how vibration learning is affected by the noise and multiple frequency components. 
Lastly, we apply our method to the real-world contact-rich task of polishing, using a hand-held power tool as shown in Fig.~\ref{fig:experiment-setup}. 
Our method can suppress the vibration and maintain a compliant profile while making contact with the work piece, in contrast to the approaches that isolate vibrations by making the robot stiffer, hence compromising safety. The results validate that the proposed method improves the performance over the original BMFLC, also in a real-world setup. 

Our paper is organised as follows: 
In Sec.~\ref{sec:litarature}, we discuss the related work in vibration suppression focusing specifically on the online learning of vibration; 
in Sec.~\ref{sec:method}, we describe our control scheme and vibration learning approach with the details of the damped BMFLC and the baseline methods;
in Sec.~\ref{sec:experiments}, we present and discuss our experiment results; and, in Sec.~\ref{sec:conclusion}, we conclude the paper with a summary of our findings.

\section{Related Work}\label{sec:litarature}

We focus on vibration suppression in applications where an external source of vibration is present, e.g., a vibrating power tool such as a drill, grinder or air chisel is used by the robot. 
This makes our problem different than the problems where the oscillations are caused by the human interaction \cite{campeau2016time}, tele-operator input \cite{yang2017personalized}, \highlight{flexible robot joints \cite{wang2017impedance,lin2023iterative}, or robot control itself  \cite{kumagai2009high}. 
In these cases, the human input or the control command can be pre-filtered to damp the resonant frequencies. 
Input shaping is an effective approach to suppress residual vibrations like these \cite{yang2024data}.}
However, in our case, the vibration is caused by external perturbation and it has to be suppressed physically. 

There are different strategies to isolate or cancel the vibration. Primarily, the methods can be divided as passive, active and semi-active \cite{yan2021bio}. Passive methods have the advantage of having low energy consumption and lower complexity, however, active methods provide more adaptability and better isolation at lower frequencies. Active methods employ various techniques such as impedance control \cite{campeau2016time,nguyen2020active}, impedance matching \cite{dai2021suppress}, adaptive filtering \cite{zou2023learning}, feedback \cite{wang2017impedance} or feedforward force control \cite{wang2014adaptive}, and trajectory optimization \cite{girgin2021optimization,tian2023vibration}. Semi-active methods offer a compromise between these two by adapting the mechanical impedance characteristics using special components \cite{liu2008semi}. 
In the case of robotics, having no additional vibration isolation hardware also has the advantages of lower payload and direct portability, that is, a control-based approach can be ported on any controllable platform. We study an active vibration suppression method with these motivation.

In the HRC literature, the vibration suppression was done either passively \cite{ciullo2021supernumerary} or in form of VIC \cite{okunev2012human,campeau2016time}. For active suppression, the amount of vibration is determined via frequency-domain \cite{okunev2012human} or time-domain analysis \cite{campeau2016time}. Then, the admittance parameters of the system are adjusted to decrease the vibration. 
In our previous work \cite{solak2023online}, we have shown the advantages of employing feedforward force control for this goal. 
The feedforward force control approach achieved better suppression in comparison to VIC approach, and also allowed keeping the impedance parameters low, which is useful in different applications. We also discussed the combination of both approaches in that work.

Our vibration suppression approach consists of learning the vibration signal online and negating it by feedforward force control. 
Vibration learning is a challenging problem that is studied for many decades. 
\highlight{We build our method on the existing online vibration learning algorithm BMFLC \cite{veluvolu2007bandlimited}.} BMFLC extends the Fourier linear combiner (FLC) algorithm \cite{vaz1989adaptive} to learn multi-frequency vibrations. 
The FLC algorithm learns a vibration of single frequency and its harmonics, which is not applicable to unknown frequency or multi-frequency vibrations. 
To answer the former limitation, the weighted-frequency Fourier linear combiner (WFLC) was proposed~\cite{riviere1998adaptive}. WFLC can track the dominant frequency of the vibration, in order to learn vibrations of unknown or drifting frequency.
BMFLC answers the latter limitation by modelling the signal as a sum of linearly separated frequency components in a selected frequency band~\cite{veluvolu2007bandlimited}. 
These approaches have the advantage of learning the vibration parameters online without the added phase lag that comes with the linear filtering approaches \cite{veluvolu2010double}. Learning the vibration as a Fourier series model allows modulating it with different time parameters, so that it can also be used as a predictor \cite{veluvolu2013multistep}. 

One limitation of BMFLC is that increasing the bandwidth size increases the computation time. \cite{veluvolu2010double} adaptively narrows the bandwidth of the BMFLC method around the relevant frequencies to increase its efficiency. However, the adaptation happens in the first few seconds of the execution and then the band is fixed. Thus, it cannot account for the future changes well. The adaptive sliding BMFLC algorithm \cite{gao2013estimation,wang2014adaptive} combines BMFLC with WFLC for a continuous adaptation of the frequency band. Another approach to deal with concept drift in vibration learning is to add a forgetting term in the learning loop \cite{atashzar2016characterization}. We also include the forgetting term in our work to accommodate the drifting frequency problem. 

The standard FLC algorithm is based on the LMS optimisation. 
Although LMS gives guarantees of convergence, it fixes the learning rate (or step-size) for linear convergence. 
Thus, many adaptive step-size methods are proposed to improve the convergence time \cite{dixit2017lms}. 
Comparison of all step-size approaches is not practically possible, however, we develop a new adaptive step-size approach for the requirements of our problem. 
We observed a limitation in the standard LMS-based learning in our preliminary results, that the learning rate cannot be increased above a certain level, otherwise it starts to learn the noise. For that reason, we added a logistic-based damping mechanism to decrease the learning rate for irrelevant frequency components as detailed in Sec.~\ref{sec:logistic-method}. 

RLS algorithm \reduce{\cite{moon2000mathematical} }and Kalman filter \reduce{\cite{burgard2005probabilistic} }are commonly used as an alternative to the LMS algorithm \cite{dixit2017lms}. We also find multiple methods in FLC-related works that employ RLS \cite{gao2013estimation,veluvolu2011estimation} or Kalman filter \cite{bo2010pathological,veluvolu2011estimation,veluvolu2013multistep,wang2011estimation}. RLS is shown to improve the estimation performance and decrease the error in the transition periods where the vibration disappears temporarily \cite{gao2013estimation}. Kalman filtering improves the estimation performance of BMFLC even further \cite{veluvolu2011estimation}. For this reason, we compared our method also with these established alternatives to the standard LMS approach.

The BMFLC algorithm was extended also in other forms. The Enhanced BMFLC algorithm \cite{atashzar2016characterization} expands the frequency band of BMFLC to include the voluntary motion to better isolate the vibration signal; and adds a forget rate to better adapt to dynamically changing frequencies. However, the former requires learning for a larger bandwidth, increasing the computational load. 
For this reason, we do not implement the former, but we use a forget rate in our method. \cite{rocon2007design} and \cite{bo2010pathological} learn the voluntary motion separately to minimise the perturbation of it. 
The quaternion WFLC algorithm \cite{adhikari2016quaternion} defines a coupling between the 3-D vibration dimensions and the grip force of the surgeon using a quaternion-based model, and shows that it decreases the estimation error. 
These enhancements could be applied together with our proposed method, however, we do not implement them as we focus on the adaptive learning rate in this work. 

A recent approach based on deep neural networks \cite{ibrahim2021real} is worth mentioning as it has shown high vibration prediction accuracy. 
However, such methods require additional data collection and training steps. Consequently, the resulting model is dependent on the selected sample population. 
Our method has the advantages of not requiring pre-training and being computationally lightweight (comparing to the ${\sim}1$ \textit{ms} run time reported in \cite{ibrahim2021real}).

The FLC-derivative algorithms are employed in many problems such as tremor cancellation for rehabilitation \cite{rocon2007design} and robotic microsurgery \cite{veluvolu2010double}, vibration suppression in laser positioning \cite{he2022multiple} and EEG estimation \cite{wang2011estimation}. Our work is the first to apply such algorithm on a high-DoF robotics task \cite{solak2023online}.



\begin{figure}[t]
\centering
  \includegraphics[width=0.98\columnwidth]{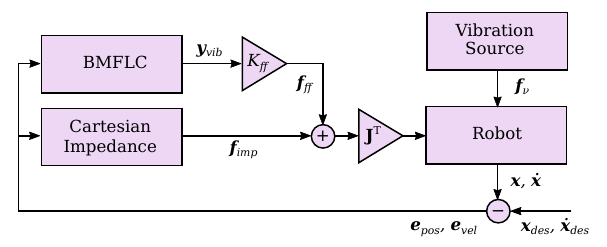} 
  \caption{Our control scheme. We aim to learn the profile of $\mathbf{f}_\nu$ and counter it by applying the feedforward force $\mathbf{f}_{ff}$. We learn the vibration model $\mathbf{y}_{vib}$ using the velocity signal $\mathbf{e}_{vel}$
  A Cartesian impedance controller is used as a base controller to achieve the desired pose $\mathbf{x}_{des}$.}  
  \label{fig:control-scheme}%
\end{figure}  

\section{Feedforward vibration suppression framework} \label{sec:method}

Our framework consists of a base controller that follows a desired trajectory and a feedforward force-based vibration suppression component. We use a Cartesian impedance controller with the BMFLC algorithm for learning and cancelling the vibration as shown in Fig.~\ref{fig:control-scheme}. The control command $\mathbf{u}$ is the sum of the Cartesian forces from the impedance controller ($\mathbf{f}_{imp}$) and the feedforward force controller ($\mathbf{f}_{ff}$) transformed to joint-space using the manipulator Jacobian $\mathbf{J}$:
\begin{equation}\label{eq:cart-force-command}%
    \mathbf{u} = \mathbf{J}^\top (\mathbf{f}_{imp} + \mathbf{f}_{ff}).
\end{equation}

The Cartesian impedance controller calculates $\mathbf{f}_{imp}$ using the trajectory following errors ($\mathbf{e}_{pos},\mathbf{e}_{vel}$): 
\begin{subequations}%
\begin{alignat}{2}%
    \mathbf{f}_{imp} &= \mathbf{K}_k \mathbf{e}_{pos} \quad+\quad \mathbf{K}_b \mathbf{e}_{vel}, \label{eq:cart-impedance}
    \\ 
    \mathbf{e}_{pos} &= \mathbf{x}_{des} - \mathbf{x},\quad \mathbf{e}_{vel} = \dot{\mathbf{x}}_{des} - \dot{\mathbf{x}}.
\end{alignat} 
\end{subequations}

The desired ($\mathbf{x}_{des}, \dot{\mathbf{x}}_{des}$) and actual ($\mathbf{x}, \dot{\mathbf{x}}$) robot poses and twists are represented in 6-D Cartesian-space. The stiffness $\mathbf{K}_k \in \mathrm{R}^{6 \times 6}$ and damping $\mathbf{K}_b$ 
$\in \mathrm{R}^{6 \times 6}$, include linear, rotational, and the coupling terms. We use $\{\mathbf{K}_k{=}400{\times}\mathbf{I}$, $\mathbf{K}_b{=}16{\times}\mathbf{I}\}$ in our experiments. The higher-order control dynamics are based on classical impedance control \cite{ott2008cartesian}. 

In our previous work, we discussed the variable impedance learning approach as part of the controller \cite{solak2023online}, however, in this study we focus on the vibration learning component. We propose an extension on the BMFLC method and compare it to other prominent extensions from the literature. Thus, the BMFLC module in Fig.~\ref{fig:control-scheme} takes different forms among the compared approaches, while rest of the framework stays the same. 
We describe the original algorithm and its extensions in the following subsection.

\subsection{BMFLC}\label{sec:bmflc-method}

The goal of the feedforward force controller is to cancel the vibration force $\mathbf{f}_\nu$ by producing the countering force $\mathbf{f}_{ff}$. We produce this force by learning the vibration signal $\mathbf{y}_{vib}$ using BMFLC and scaling it by the gain $K_{ff}$:
\begin{equation}\label{eq:bmflc-force}%
    \mathbf{f}_{ff} = K_{ff} \mathbf{y}_{vib}.
\end{equation}

The vibration is modeled as a Fourier series model. We formulate an arbitrary Cartesian dimension of $\mathbf{y}_{vib}$ as $y_{vib}$:
\begin{subequations}%
\begin{alignat}{2}%
    y_{vib}(t) &= \sum^{L}_{r=0} w_r(i) g_{r}(t) + w_{(r+L)}(i) g_{(r+L)}(t),\label{eq:fourier-series}
    \\ 
    g_{r}(t) &= sin(2\pi \nu_r t),\\
    g_{(r+L)}(t) &= cos (2\pi \nu_r t).
\end{alignat} 
\end{subequations}

The model contains a pair of basis functions ($g_r(t), g_{(r+L)}(t)$) for each frequency component $\nu_r$. The linear combination of these basis functions produce a different vibration signal depending on the choice of $w_k$. We use $t$ to indicate time and $i$ to indicate a discrete iteration step. We drop $i$ in our text whenever it is redundant.

In BMFLC, we split a target frequency band $[a_\nu,b_\nu)$ into $L$ equal intervals and assign a pair of weights and basis functions to each delimiting value $\nu_r$:
\begin{equation}
    \nu_r = a_\nu + r(|b_\nu-a_\nu|/L), \quad r \in [1,L).
\end{equation}


The target vibration is learned by minimising the error $e(t)= |f_\nu(t) - y_{vib}(t)|$. In our system, we use velocity error $e_{vel}$ directly as $e(t)$, because the output is already subtracted from the velocity in the form of feedforward control. Each weight is optimised iteratively by the LMS algorithm \cite{vaz1989adaptive}:
\begin{equation} \label{eq:bmflc-update}
    w_k(i+1) = w_k(i) + 2\eta g_k(t) e(t) , \quad k \in [1,2L).
\end{equation}

Originally, a constant step-size (or learning rate) $\mu$ is used for the weight updates. The choice of $\mu$ determines the speed of adaptation, however, for high values of this parameter the algorithm can become unstable. In vibration learning context, it starts to learn the noise.

For this reason, we propose a new adaptive step-size approach that changes the learning rate depending on the current value of a weight $w_k$. Details of our method is described in Sec.~\ref{sec:logistic-method}. We compare our modification to \textit{RLS-based} and \textit{Kalman-based} optimisation approaches, thus we also describe these methods in Sections~\ref{sec:rls-method} and \ref{sec:kalman-method}, respectively.

Before presenting the adaptive step-size approaches, we rewrite \eqref{eq:bmflc-update} with a step-size function $\mu(t)$ which will be re-defined by the compared methods:
\begin{subequations}%
\begin{alignat}{2}%
    w_k(i+1) &= w_k(i) \lambda + \mu(t) e(t),
    \\ 
    \mu(t) &= 2 \eta g_k(t).
\end{alignat} 
\end{subequations}

\begin{figure}[t]
\centering
  \includegraphics[width=0.98\columnwidth]{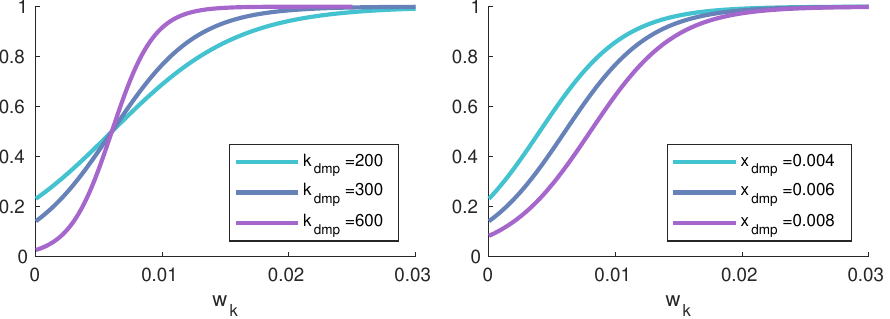} 
  \caption{$\mu_{dmp}(t)$ maps lower weights to damped learning rates to reduce noise in learning. The function is plotted for $\eta g_k(t){=}1$ to demonstrate the effects of parameters $k_{dmp}$ and $x_{dmp}$. }  
  \label{fig:logistic-fcn}%
\end{figure}  

We also added a forgetting rate $\lambda$ to better account for the drifting effects as described in \cite{atashzar2016characterization}. This is usually a value less than but close to $1$. Smaller values cause faster forgetting, or in other words a smaller window size.

\subsubsection{Damped BMFLC}\label{sec:logistic-method} 
The LMS algorithm follows the gradient of the error function with constant rate $\mu$. Increasing $\mu$ means faster convergence, however, we cannot increase it too much because it starts learning the noise. To learn the relevant components faster while disregarding the irrelevant components we add a damping mechanism in form of logistic function. An indicator that helps us to distinguish the noise from useful frequencies is the learned weight $w_k$, which will be lower for the irrelevant components. Therefore, we damp the step-size function for lower values of $w_k$ as follows:
\begin{equation}
    \mu_{dmp}(t) = \frac{\eta g_k(t)}{1+e^{-k_{dmp} (w_k - x_{dmp})}}. 
\end{equation}

The constants $k_{dmp}$ and $x_{dmp}$ determine the steepness and the midpoint of the sigmoid shape that the logistic function defines. $\mu_{dmp}(t)$ and the effects of its parameters are illustrated in Fig~\ref{fig:logistic-fcn}. The use of logistic function prevents the step-size from growing unstably. As seen on the figure, $\mu_{dmp}(t)$ produces lower step-sizes for small $w_k$ and it approaches to the original learning rate $\mu$ for larger $w_k$. Although our modification decreases the overall learning speed for a given $\mu$, it allows us to choose larger $\mu$ values as it is harder to learn the noise with it. Thus, it can adapt faster than the constant step-size approach.


\subsubsection{Recursive least squares}\label{sec:rls-method} 
We implement the RLS \reduce{\cite{moon2000mathematical} }step-size function as described in \cite{veluvolu2011estimation}:
\begin{equation}\label{eq:rls-stepsize}
    \pmb{\mu}_{rls}(t) = \frac{\mathbf{P}(i) \mathbf{g}(t)}{\lambda_{rls} + \mathbf{g}(t)^\top\mathbf{P}(i)\mathbf{g}(t)}.
\end{equation}

We use the bold notation here for vector form containing all components for $k \in [1,2L)$. $\lambda_{rls}$ is the forgetting factor and $\mathbf{P}(i)$ is the covariance matrix\footnote{{\label{matrix-size}$\mathbf{P}(i), \mathbf{Q}_{kf} \in R^{2L\times 2L}$}}, updated at each iteration $i$:
\begin{equation}\label{eq:rls-p}
    \mathbf{P}(i+1) = \frac{1}{\lambda_{rls}} [\mathbf{P}(i) - \pmb{\mu}_{rls}(t) \mathbf{g}(t)^\top \mathbf{P}(i)].
\end{equation}

\subsubsection{Kalman filtering}\label{sec:kalman-method}
The Kalman filter \reduce{\cite{burgard2005probabilistic} }is implemented with random walk model following \cite{veluvolu2011estimation}. The measurement noise and state noise are assumed to be uncorrelated and zero mean Gaussian processes, with covariance parameters $R_{kf}$ and $\mathbf{Q}_{kf}$\textsuperscript{\ref{matrix-size}}. Kalman-based step-size function is computed as:
\begin{equation}\label{eq:kalman-stepsize}
    \pmb{\mu}_{kf}(t) = \mathbf{P}(i) \mathbf{g}(t)^\top (\mathbf{g}(t)^\top\mathbf{P}(i) \mathbf{g}(t)+R_{kf})^{-1}.
\end{equation}

The state error covariance $\mathbf{P}(i)$ is updated at each iteration $i$:
\begin{equation}\label{eq:kalman-p}
    \mathbf{P}(i+1) = [\mathbf{I} - \pmb{\mu}_{kf}(t) \mathbf{g}(t)] \mathbf{P}(i) + \mathbf{Q}_{kf}
\end{equation}

\section{Experiments}\label{sec:experiments}

Our experiments aim to find evidence on multiple questions: 
\begin{enumerate}[label=\textbf{Q\arabic*.}, ref=\textbf{Q\arabic*}]
    \item\label{q:performance} Does the damped BMFLC method improve the vibration learning performance? 
    \item\label{q:param-sensitivity} Is it robust to parameter changes?
    \item\label{q:multi-freq} Can it suppress time-varying multi-frequency vibrations?
    \item\label{q:time-cost} Is it computationally efficient to run in high-frequency?
    \item\label{q:limits} What are the limits of the method under different vibration and noise conditions?
    \item\label{q:real-robot} Can it be used as part of a real-world physical robot interaction controller?
\end{enumerate}
We carry out extensive simulation experiments to answer the first five questions. We also carry out real-robot experiments to investigate the last question. A supplementary video of the experiments is available online. The question numbers are referred in the text to facilitate reading.

\subsection{Simulation experiments}

We simulate a mass-spring-damper (MSD) system in Matlab Simulink as our testbed (\textit{mass: 3.6, stiffness: 400, damping: 100}). The mass position is controlled using an impedance controller \eqref{eq:cart-impedance} to follow a reference trajectory $\mathbf{x}_{r}$. The system is continuously perturbed by time-varying multi-frequency vibration force $\mathbf{f}_\nu$ and a normally distributed noise $\mathbf{f}_n$. This type of vibration is applied in all experiments to test for \ref{q:multi-freq}.  Both the reference trajectory and vibration are generated randomly as described in Sec. \ref{sec:synthesis}.

First, we compare the damped BMFLC method to different stepsize approaches that are described in Sec. \ref{sec:method}: \textit{Original}, \textit{RLS-based} and \textit{Kalman-based}. The details of this experiment is discussed in Sec. \ref{sec:sim-compare}. Then, we apply our proposed method under different vibration and noise conditions to assess the limits of our method (\ref{q:limits}), as detailed in Sec. \ref{sec:sim-limits}.

Our performance criteria evaluate the vibration \textit{suppression rate (SR)} (\ref{q:performance}) and the \textit{computation time} of each adaptive step-size iteration (\ref{q:time-cost}). We calculate the SR as the percentage of reduction of the mean squared vibration amplitude:

\begin{equation}
   \text{SR} = 1 - \frac{\sum^{}_i (\mathbf{f}_\nu(i)-\mathbf{f}_{ff}(i))^2}{\sum^{}_i \mathbf{f}_\nu(i)^2}
\end{equation}

\subsubsection{Synthetic data}\label{sec:synthesis}

We procedurally generate the voluntary motion trajectory $\mathbf{x}_{r}$ and vibration force $\mathbf{f}_\nu$ using a Fourier series model \eqref{eq:fourier-series}, i.e., each motion aggregates multiple sinusoidal waves of varying frequency $\nu$ and phase angle $\phi$. 

The frequency $\nu_k$ and phase $\phi_k$ of each component $k$, and the number of components $N_\nu$ are sampled from uniform distributions of $\mathcal{U}(a_{\nu},b_{\nu})$, $\mathcal{U}(a_{\phi},b_{\phi})$, $\mathcal{U}(a_{N},b_{N})$, respectively. 
In case of the voluntary motion $\mathbf{x}_r$, an exponential distribution is used to prioritise lower frequencies: $\nu_k \sim \textit{Exp}((b_\nu{-}a_\nu)/5) {+} a_\nu$. 
The amplitude $\xi_k$ of each component $k$ is sampled from a normal distribution $\mathcal{N}(m_{\xi_k}, s_{\xi_k})$ with means $m_{\xi_k}= (\xi_{total} / b_N)  (b_N{-}k)$, i.e., linearly decreasing values to obtain an expected total amplitude of $\xi_{total}$ and to create a hierarchy between the frequency components. $b_N$ is the upper limit of the number of components ($N_\nu$).

\begin{figure}[t]
\centering
  \begin{tabular}{c}
    \includegraphics[width=0.95\columnwidth]{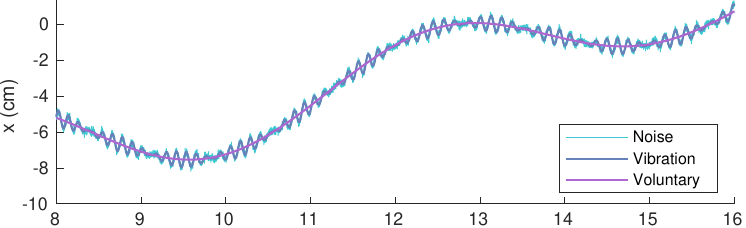} 
    \\             
    \includegraphics[width=0.98\columnwidth]
    {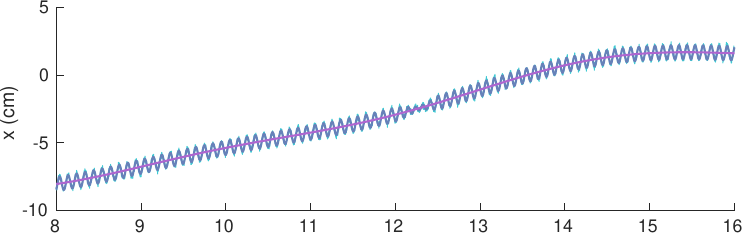} 
    \\             
    \includegraphics[width=0.95\columnwidth]
    {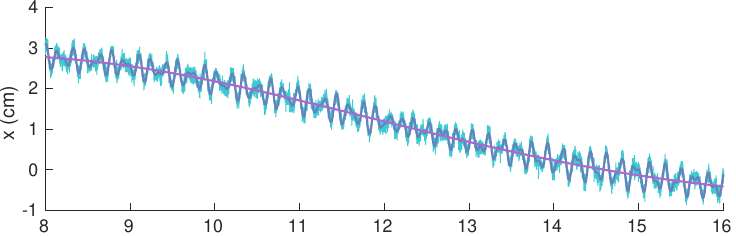} 
    \\             
    \includegraphics[width=0.96\columnwidth]{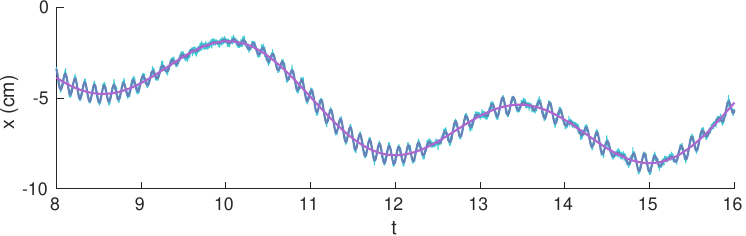} 
  \end{tabular}
  \caption{Selected examples of synthesised vibrating motions. The motions have different number of vibration components with randomised frequency and amplitude characteristics. Each of these motions could describe a daily activity such as reaching or waving of a hand with noticeable tremor. The frequency drift happens at $12.25$ \textit{s}.}  
  \label{fig:synthetic-motions}%
\end{figure}  

After creating the motions with multi-frequency vibrations, we also create a \emph{frequency drift}, i.e., a gradual change in the frequencies of the vibration components. The drift happens at a predefined time point ($12.25$ \textit{s}), lasting for a predefined duration ($0.5$ \textit{s}). The new frequency $\nu_k'$ after the drift is sampled from a normal distribution $\mathcal{N}(\nu_k, s_{\nu})$, centered at the initial frequency $\nu_k$. The transition happens smoothly as the amplitude of the old frequency component decreases to zero linearly while that of the new component increases to $\xi_k$.

The final motion data is obtained as the sum of the voluntary position trajectory $\mathbf{x}_r$, the vibration $\mathbf{f}_\nu$ and a Gaussian noise~$\mathbf{f}_n \sim \mathcal{N}(0, s_{n})$.
Some of the motions obtained using this procedure\footnote{We use the  parameters $\{a_\nu{:}6$, $b_\nu{:}10$, $a_N{:}1$, $b_N{:}3$, $\xi_{total}{:}0.6$, $s_{\xi_k}{:}0.05/N_\nu$, $s_{\nu}{:}0.5\}$ for vibration, $\{a_\nu{:}0.01$, $b_\nu{:}0.3$, $a_N{:}7$, $b_N{:}10$, $\xi_{total}{:}10$, $s_{\xi_k}{:}0.3/N_\nu\}$ for voluntary motion and $\{a_\phi{:}0$, $b_\phi{:}2\pi$, $s_n:0.001\}$ commonly as default in the experiments.} are depicted in Fig. \ref{fig:synthetic-motions}.

\begin{figure}[t]
\setlength\tabcolsep{0.5pt}
\centering
  \includegraphics[width=0.98\columnwidth]{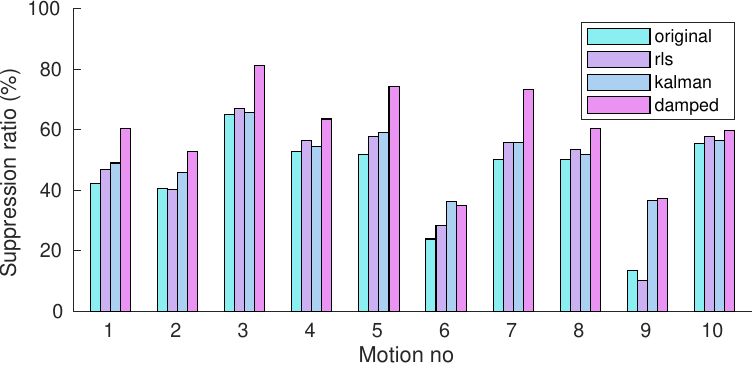} 
  \caption{The suppression rates of the compared approaches on different synthesised motions in simulation experiments. The motions differ by the number and characteristics of frequency components that are randomly generated as described in \ref{sec:synthesis}. The proposed method achieves the best or near-best performance on all motions.}  
  \label{fig:comparison-results}%
\end{figure}  

\begin{figure}[t]
\centering
    \includegraphics[width=0.95\columnwidth]{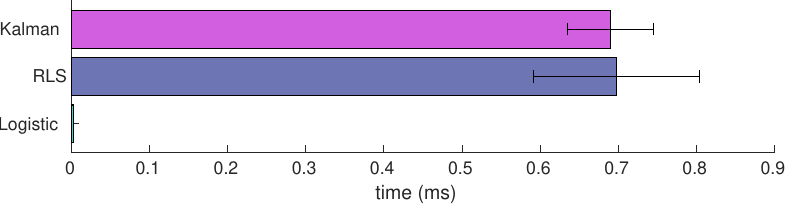}   
  \caption{Run times of the adaptive step-size methods in Matlab, averaged over 100 repetitions for $L{=}120$.}  
  \label{fig:stepsize-timeit}%
\end{figure}  

\subsubsection{Method comparison}\label{sec:sim-compare}

Each experiment case is repeated on a set of $10$ randomly generated vibrating motions ($\mathbf{x}_{r},\mathbf{f}_\nu$). 

We use the \texttt{fminsearch} function of Matlab to find the optimal stepsize parameters for each method (for fixed $L{:}100$, $\lambda{:}0.9999$). Optimal algorithm parameters are different for each generated motion due to different vibration components ($\nu_k,\xi_k$) present. However, in real-world applications, it is not always feasible to optimise these parameters for each task. Thus, the method should produce good results using the same parameter set for different motions (\ref{q:param-sensitivity}). In order to avoid over-fitting, we find the optimal parameters for the first $5$ motions and assign the average value of the $5$ values as the general parameter set. The experiments are repeated for all $10$ motions using the same parameter set which is determined this way.

The suppression results are presented in Fig.~\ref{fig:comparison-results}. The damped BMFLC method achieves significantly better suppression performance for majority of the motions and matches the highest-performing approach in the remaining few, providing a strong evidence for \ref{q:performance}. These results are obtained with the general parameter sets on time-varying multi-frequency vibrations, supporting that \ref{q:param-sensitivity} and \ref{q:multi-freq} are satisfied. The performance improvement can be attributed to the ability of our method to enable high learning rates, without falling into the trap of learning the noise. Our method achieves this through a simpler mechanism in comparison to the \textit{RLS-based} and \textit{Kalman-based} step-size approaches, thus obtain a faster adaptation.

We calculate the step-size computation times in Matlab using the \texttt{timeit} function. Fig~\ref{fig:stepsize-timeit} shows the mean and standard deviation of 100 repetitions. These results affirm that our method has a major advantage in computation time and it is adequate to run in real-time (\ref{q:time-cost}). 
The matrix operations required for \textit{RLS-based} and \textit{Kalman-based} step-size approaches render them more expensive to compute, to the point that our Matlab simulation cannot be run in real-time on our hardware (Ubuntu 20.04, Intel Core$^{\text{TM}}$ i7 2.3 GHz x 16, 15.4 GB RAM) when using these methods. 

\begin{figure}[t]
\centering
  \begin{tabular}{c}
    \includegraphics[width=0.97\columnwidth]{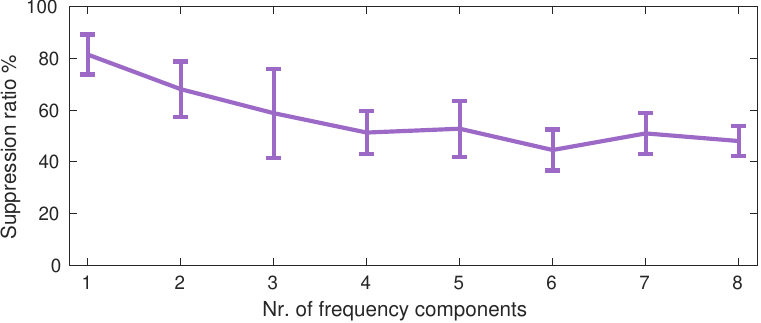} 
    \\             
    \includegraphics[width=0.97\columnwidth]{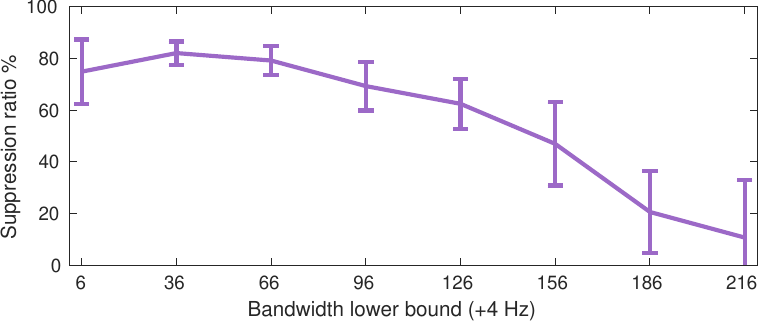} 
    \\             
    \includegraphics[width=0.97\columnwidth]{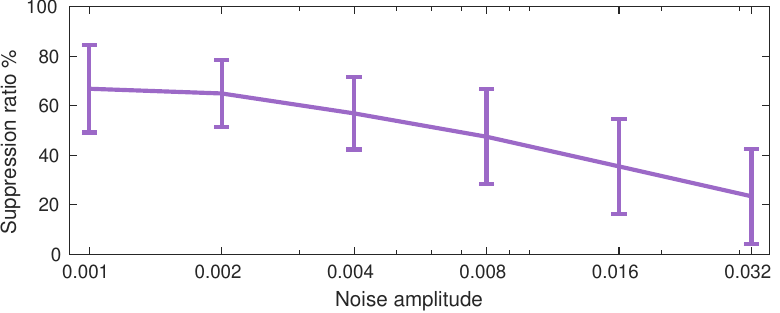} 
  \end{tabular}
  \caption{Results of the limit analysis experiments. The plots show the change of vibration suppression rate with changing experiment conditions. 
  The number of frequency components in the vibration signal, the range of vibration frequencies and the noise amplitude are shown to be important limiting factors for active vibration suppression.}  
  \label{fig:limit-analysis}%
\end{figure}  

\subsubsection{Limit analysis}\label{sec:sim-limits}
In order to answer \ref{q:limits}, we examine the limits of our method under higher number of frequency components~$N_\nu$, larger vibration frequencies~$(a_\nu, b_\nu)$ and larger noise amplitudes~$s_n$. 
We increase values of these parameters and repeat each case $10$ times to account for variability. 
We re-synthesise the motions with these parameters as described in Sec.~\ref{sec:synthesis}. We use the optimal stepsize parameters for each case.
The results are presented in Fig.~\ref{fig:limit-analysis}:

\begin{itemize}
    \item $N_\nu$: The learning performance is inversely related to the complexity of the vibration signal. Adding each of the first three frequency components visibly decrease the suppression rate. After the $3^{rd}$ component, the performance converges to a certain level. This may be because of that the effect of minor components approaches zero. 
    Please note that $N_\nu$ is the number of actual frequency components in the target signal. Our algorithm does not impose any explicit limit on the number of learned frequency components, and aims to learn as many as possible. 
    
    \item $(a_\nu, b_\nu)$: We show the lower bound of the frequency band $a_\nu$ on the figure for simplicity. $b_\nu$ is always set to be $a_\nu+4$. We can see that the performance does not change significantly until approaching to ${\sim}100$ \textit{Hz}. The performance declines significantly above this frequency. 
    We attribute this decline to the decrease of sampling resolution of the vibration signal as the vibration frequency gets closer to the sampling frequency, which is $1000$ \textit{Hz} in our work. 

    \item $s_n$: We fixed the vibration amplitude ($\xi_{total}{=}0.5$) and increased the noise amplitude exponentially. 
    The results show that the suppression rate is \textit{inversely logarithmically} proportional to the noise amplitude.
    One of the main factors that influence the performance is the noise to signal ratio. 
    Because when the noise amplitude increases, it will activate random BMFLC weights. 
    For the same reason, we cannot set the learning rate $\mu$ too high, otherwise it starts to learn the noise. 
\end{itemize}

\subsection{Real-world experiments}\label{sec:real-world-exp}

We assess our method also on a real-world polishing task to test for \ref{q:real-robot}. Our setup consists of a 7-DoF Franka Emika Panda robot arm carrying a hand-held drill with polisher head (Fig.~\ref{fig:experiment-setup}). The drill is modified to be controlled via an Arduino-USB interface. The drill and the robot are controlled simultaneously using ROS programs, so that the experiment conditions are reproduced without manual intervention. \highlight{We run ROS on Ubuntu with a real-time Linux kernel as it is also required by the Franka libraries. This allows us to answer the requirements of vibration suppression by running our controller at 1000 \textit{Hz}.}

\begin{figure}[t]
\centering
    \includegraphics[width=0.98\columnwidth]{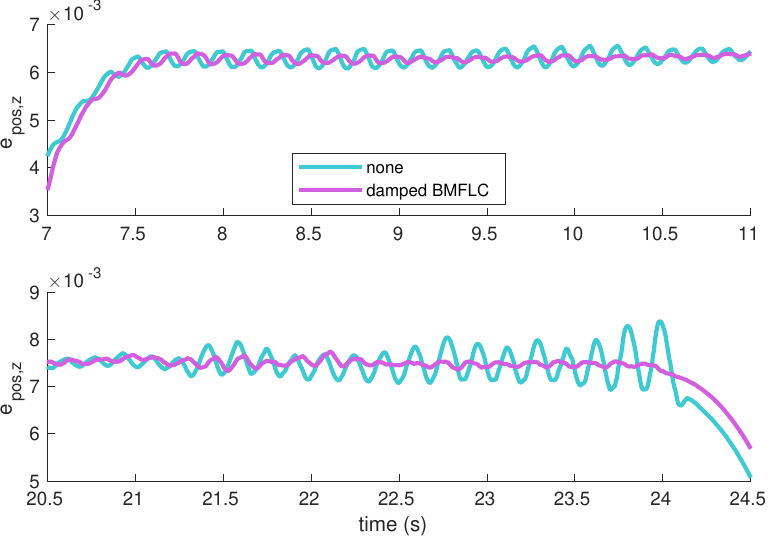}  
  \caption{ Time series of the end-effector position error in \textit{z}-axis ($e_{pos,z}$) during a polishing experiment with drill mode \textit{100}. 
  The top plot shows the initial contact and the bottom plot shows the end of the experiment. 
  The damped BMFLC method visibly decreases the positional vibration in comparison to the case without suppression.
  }  
  \label{fig:polishing-timeseries}%
\end{figure}

In each experiment, the end-effector makes contact with the surface, polishes the wooden block while moving linearly for \textit{20 cm} in \textit{16 s} until it breaks contact with the surface as shown in Fig.~\ref{fig:experiment-setup}. The vibrations are noticeably amplified during physical contact with the surface as seen in Fig.~\ref{fig:polishing-timeseries}.

We apply our online vibration suppression method to reduce the vibration in real-time. We evaluate our method with different drill modes (\textit{80, 100, 120}) of increasing power which produce different vibration frequencies (\textit{5.5, 6.3, 8 Hz}). We compare the vibration reduction performance to the baseline of constant step-size. We first explore different $\mu$ to choose the optimal value for each method (Fig.~\ref{fig:error-vs-mu-realworld}), on each drill speed. Then, we repeat the experiment $5$ times for each case: 1) no suppression; 2) original BMFLC; 3) damped BMFLC.\footnote{The other parameters are $\{L{:}240$, $\lambda{:}0.99999$, $x_{dmp}{:}0.009$, $k_{dmp}{:}350\}$.} 


\begin{figure}[t]
\centering
  \includegraphics[width=0.98\columnwidth]{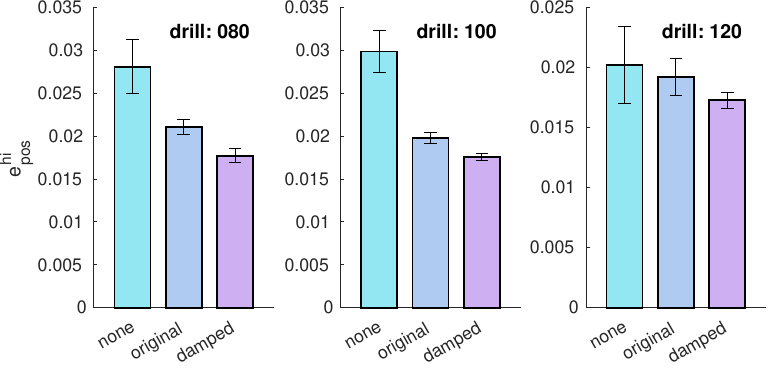} 
  \caption{Real-world experiment results for different drill modes. The bars indicate the mean error for 5 repetitions. The damped BMFLC improves the performance consistently.}  
  \label{fig:realworld-results}%
\end{figure}  

\subsubsection{Results and Discussion} 
We quantify the vibration as the mean squared error (MSE) of the band-pass filtered (\textit{3-100 Hz}) position error over the time interval of the experiment, noted as $e_{pos}^{hi}$. Filtering removes the effect of the trajectory following error (${<}3$ \textit{Hz}) and indicates only the vibration.

\begin{figure}[t]
\centering
  \includegraphics[width=0.98\columnwidth]{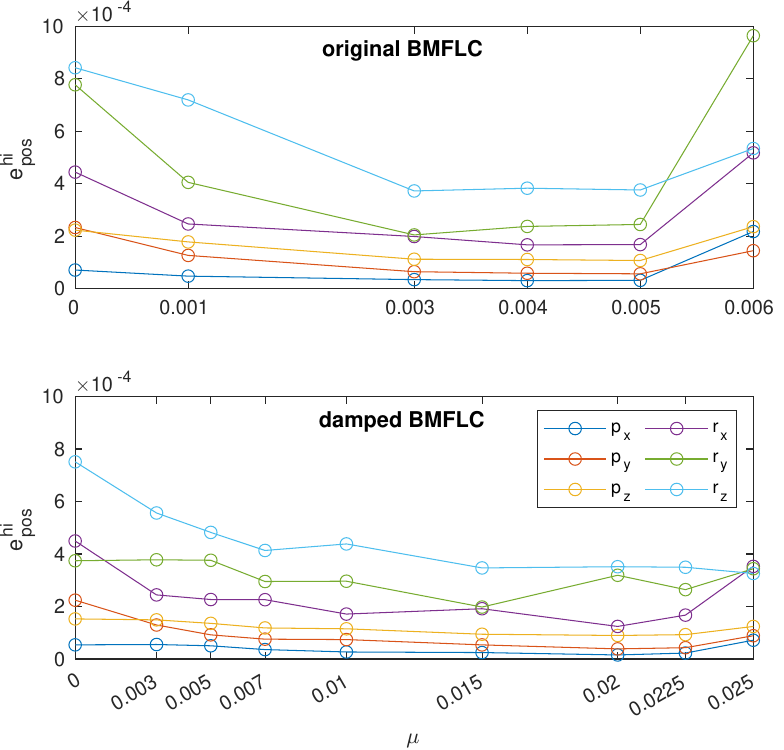} 
  \caption{The change of vibration error $e^{hi}_{pos}$ in different dimensions w.r.t. the learning rate $\mu$ for drill mode \textit{100} in polishing experiment. The scale of $\mu$ changes between the approaches. The error starts increasing for too large $\mu$ due to learning the noise. }  
  \label{fig:error-vs-mu-realworld}%
\end{figure}  

The outcomes of different methods are presented side-by-side in Fig.~\ref{fig:realworld-results}, under different drill speeds (or vibration frequencies). The damped BMFLC method consistently achieves better performance than the original BMFLC method. These results validate the advantage and applicability of our method in real-world conditions (\ref{q:real-robot}). The position error plot shows the visible reduction in vibration with our method (Fig.~\ref{fig:polishing-timeseries}). Because the method learns the vibration online, the difference increases gradually and becomes more pronounced by the end.

The difference of the suppression seems smaller for the drill mode \textit{120} in Fig.~\ref{fig:realworld-results}. However, careful examination reveals that the vibration level $e^{hi}_{pos}$ is brought down to a similar range as in the other cases for both \textit{original} (${\sim}0.0200$) and \textit{damped BMFLC} (${\sim}0.0175$). The difference is smaller mainly because the original vibration is not as strong as in the other cases.

A limiting factor in these experiments is the absence of a specific sensor for directly observing the vibration. We use the velocity signal, measured by the robot encoders, to quantify and model the vibration. Attaching an accelerator to the end-effector could provide a clearer vibration signal and further improve the learning performance. 

\begin{figure}[t]
\centering
  \includegraphics[width=0.98\columnwidth]{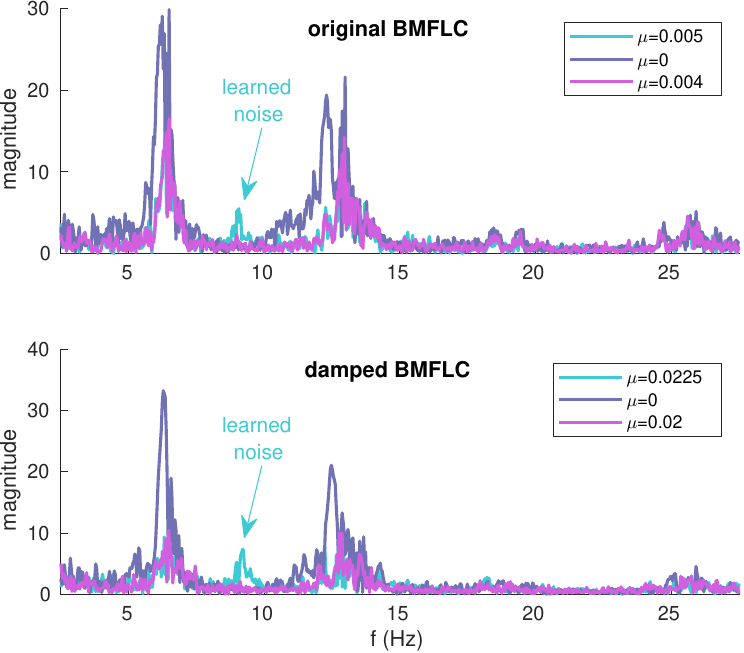} 
  \caption{Frequency magnitude plots for selected $\mu$ values for the same experiment as in Fig.~\ref{fig:error-vs-mu-realworld}. These plots are obtained by applying Fourier transformation on the \textit{y}-axis of velocity signals. The cyan signals demonstrate the effect of learning noise in both cases when $\mu$ is too high. }  
  \label{fig:learning-noise-freq}%
\end{figure}  

We also plot $e^{hi}_{pos}$ for all 6 Cartesian dimensions in Fig.~\ref{fig:error-vs-mu-realworld}. This plot shows how the vibration decreases for increasing learning rates ($\mu$) for step-size approaches with drill mode \textit{100}. We actively suppress vibration in 3 dimensions, but we observe that the vibration decreases in all 6 dimensions including the orientation. The plot also shows the upper limit of $\mu$: the vibration error starts increasing after $\mu{>}0.004$ for \textit{original} and $\mu{>}0.020$ for \textit{damped BMFLC}, because the algorithm gets too sensitive and starts learning the noise. 

We can clearly see the effect of learning the noise on the frequency magnitude plots. We transformed the velocity signal of selected runs into frequency domain and plotted them in Fig.~\ref{fig:learning-noise-freq}. We see that with the preferred learning rate $\mu{=}0.020$, the damped BMFLC can learn the correct target frequency and reduce its magnitude. 
Although we run the BMFLC with a target band of \textit{3-9 Hz}, we see a reduction in the harmonics of the dominant frequency of $6.3$ \textit{Hz}: $12.6, 18.9$... 
However, if we further increase the learning rate ($\mu{=}0.0225$), it starts learning the noise. The controller introduces the learned noise frequency in the motion. In this case, frequency of the learned noise is $9$ \textit{Hz}, which is the upper limit of the target frequency band. For the original BMFLC approach, the method starts learning the noise already at $\mu{=}0.005$, because it lacks a noise reduction mechanism.

The effect of damped BMFLC is highlighted by the noise-learning behaviour. The logistic function (Fig.~\ref{fig:logistic-fcn}) damps lower weights which belong to noisy components, and consequently encourages learning more relevant frequency components. This allows us to choose larger $\mu$ values ($0.020$) in comparison to the original BMFLC ($0.004$), without learning the noise. Larger learning rates result in quicker response and higher vibration reduction. 

\section{Conclusion} \label{sec:conclusion}

This work extends our study on active vibration cancellation in physical interaction tasks with collaborative robots. 
The logistic function based damping mechanism increases robustness to noise and consequently enables faster learning rates.
Extensive simulation experiments validate that the damped BMFLC performs consistently better than the original BMFLC and it is superior to the RLS and Kalman-based approaches in means of suppression performance and computation time.  
The experiments also confirm that the method is fairly robust to parameter changes.
It can deal with time-varying multi-frequency vibrations which require high adaptability and flexibility. 

Our experiments with more extreme conditions of vibration frequency and noise presents a reference on the limitations of the method for the future work. It also informs us that noise and frequency conditions affect algorithm performance significantly, thus it requires attention when generalising the reported results of a particular paper.

The robotic polishing experiments validated the applicability of our method on a real-world physical interaction task. It achieved better performance than the original BMFLC method, significantly suppressing the vibration in the supported frequency ranges.


\backmatter


\bmhead*{Funding} This work was supported by the European Union Horizon Project TORNADO (GA 101189557).
\bmhead*{Data availability} The data collected in the real-world experiments is publicly available at \href{https://doi.org/10.6084/m9.figshare.29828711.v3}{https://doi.org/10.6084/m9.figshare.29828711.v3}.



\addtolength{\textheight}{-4.5cm} 

\bibliography{main2025}

\end{document}